\documentclass[runningheads]{llncs}
\usepackage[T1]{fontenc}
\usepackage{graphicx}
\usepackage{hyperref}
\usepackage{color}

\usepackage[dvipsnames]{xcolor}

\usepackage{amsmath,amssymb}

\begin{document}
\title{Closing the Loop on the Poppy Humanoid: Bipedal Locomotion with Linear-Quadratic Control and Learned Cost Functions}
\titlerunning{Closed-Loop Control of the Poppy Humanoid with Learned Cost Functions}
%
\author{
Xulin Chen\inst{1}\orcidID{0009-0004-9140-6547} \and
Borui He\inst{1}\orcidID{0009-0002-2164-8466} \and
Ruipeng Liu\inst{1}\orcidID{0009-0006-4353-4952} \and
Naveed Tahir\inst{1}\orcidID{0009-0002-5078-8927} \and
Zhenyu Gan\inst{1}\orcidID{0000-0002-5972-9600} \and
Garrett E. Katz\inst{1}\orcidID{0000-0002-5036-8394}
}
\authorrunning{X. Chen et al.}
%
\institute{
Syracuse University, Syracuse NY 13244, USA\\
\email{\{xchen168,bhe100,rliu02,ntahir,zgan02,gkatz01\}@syr.edu}\\
Contact Author: Garrett E. Katz
}
\maketitle              
\begin{abstract}
The Poppy Humanoid is an open-source, low-cost robot suitable for research and education in artificial intelligence.  However, we are unaware of any published methodology that achieves reliable, unassisted bipedal locomotion on the standard Poppy hardware.  This paper contributes a functional closed-loop walking controller for Poppy, based on the linear-quadratic regulator (LQR) framework for trajectory tracking.  Starting with data collected from open-loop playback of a nominal walking trajectory, our proposed method learns a quadratic cost function for an LQR controller that substantially improves the reliability of the motion.  The closed-loop controller is validated empirically, demonstrating statistically significant improvements in walking performance compared to open-loop trajectory playback.


\begin{center}
\textbf{Regular Research Paper}
\end{center}

\end{abstract}
\section{Introduction}

\begin{figure}[t]
  \centering
  \includegraphics[width=.28\columnwidth]{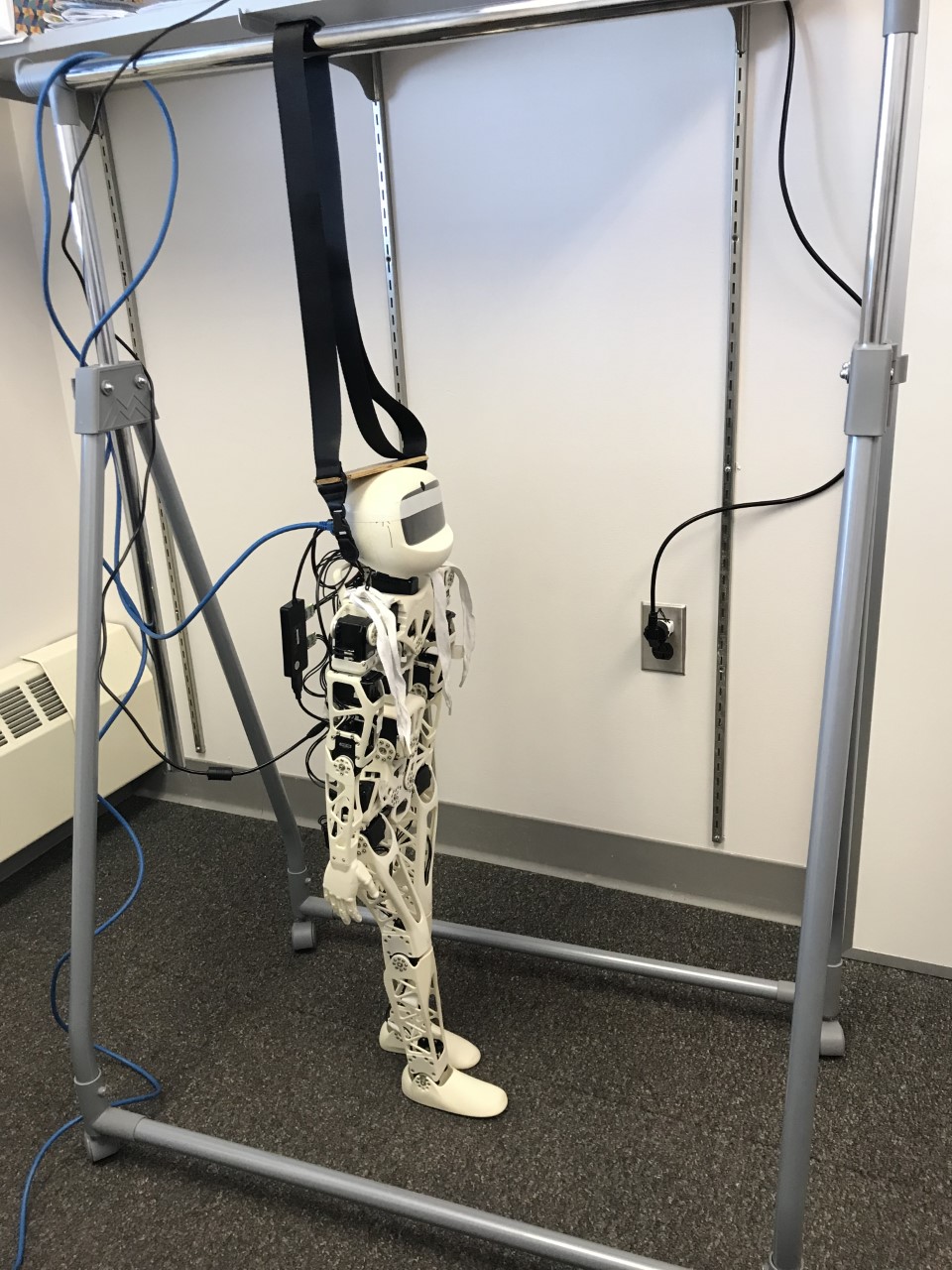}
  \includegraphics[width=.217\columnwidth]{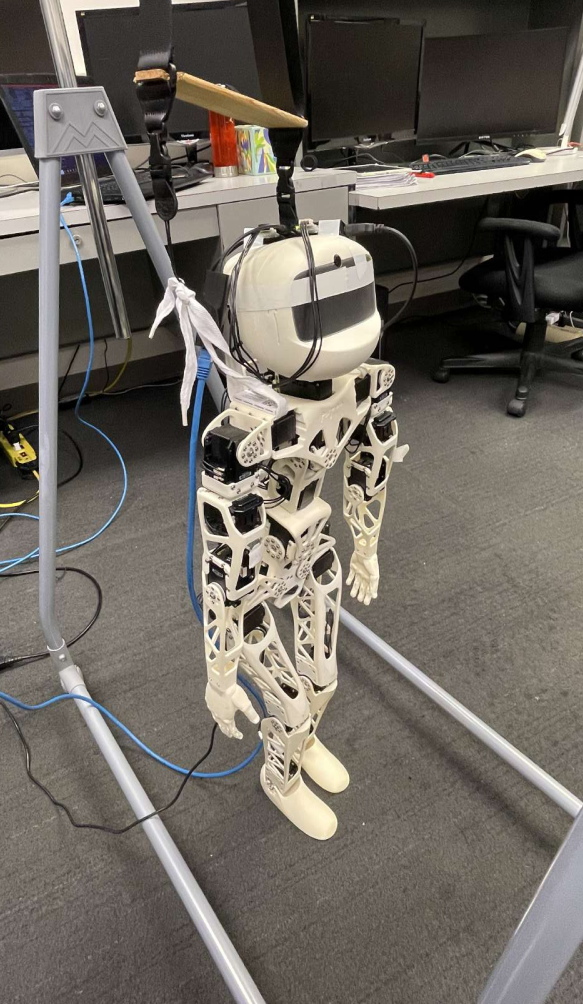}
  \includegraphics[width=.219\columnwidth]{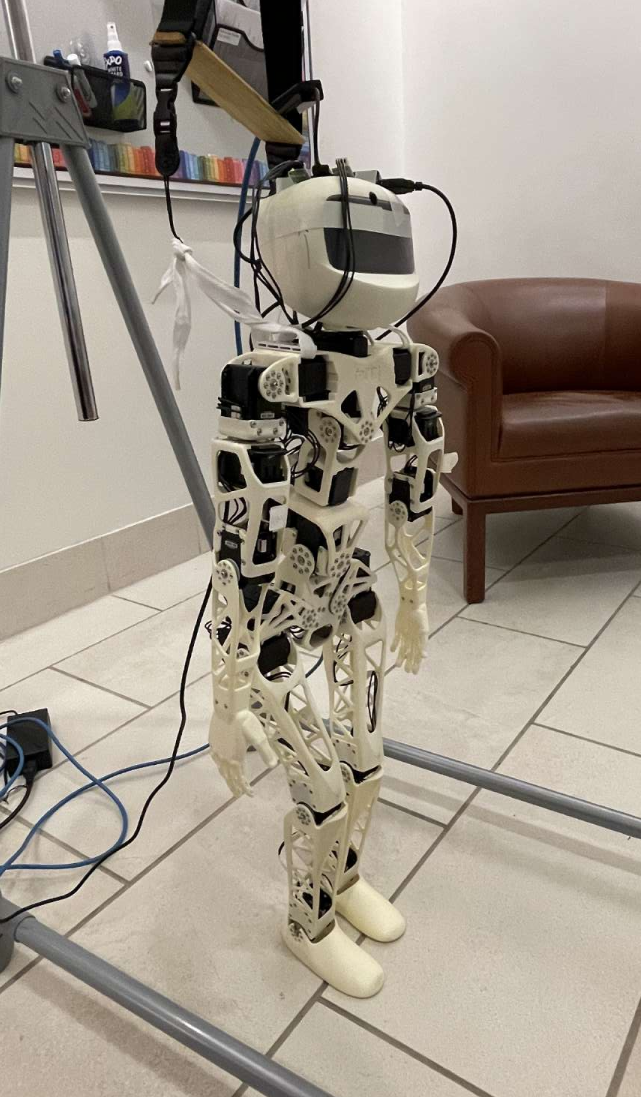}
  \includegraphics[width=.217\columnwidth]{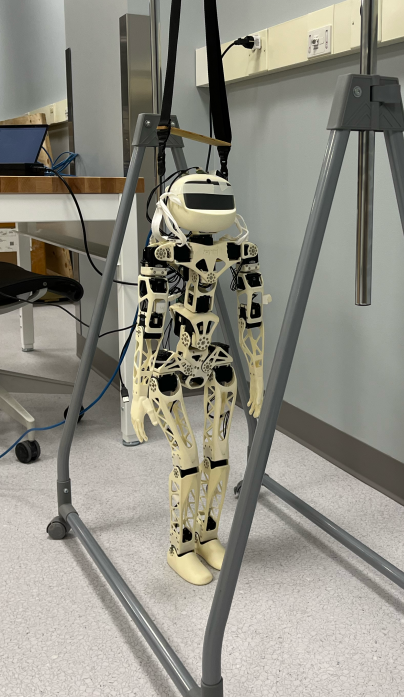}
  \caption{The Poppy hardware and four locations used in our experiments.  To prevent hardware damage during falls, we fashioned an experimental harness from a consumer clothing rack and camera strap, with a custom wooden insert to avoid pressure on the sides of the head when the strap is taut.  We only count footsteps as ``successful'' if the strap is slack for the entirety of the motion without any human intervention or assistance.}
  \label{fig:poppy}
\end{figure}

The Poppy Humanoid is an open-source, child-sized robot, developed roughly 13 years ago for research and education \cite{lapeyre2013poppy}.  Aside from the electronics and motors, the robot limbs are all 3D-printed, resulting in a relatively light-weight, low-cost, and extensible robotic platform. Poppy has been used for research in bipedal motion \cite{lapeyre2013thigh} and human-machine interaction \cite{erol2019toward}.  The Poppy hardware and experimental harness in our lab is shown in Fig.\@ \ref{fig:poppy}.

Poppy's low cost is beneficial for researchers and educators, but also incurs several limitations that hinder reliable bipedal locomotion.  The motors only support position control, and the maximum achievable control rate is relatively low (well below 100Hz).  Furthermore, in the standard off-the-shelf design (before any customization), the foot contact surfaces are relatively small and there is only one rotational axis at each ankle.  This limits inherent controllability of the motion.  Lastly, the standard hardware package available for purchase does not include foot pressure sensors or inertial measurement units, so the robot state is only partially observable and difficult to estimate.

Perhaps due to these hardware challenges, to the best of our knowledge, existing literature on the Poppy Humanoid does not provide any effective technique for autonomous walking.  Past bipedal walking studies with Poppy have required a trained expert user to continually hold Poppy's hands to help it keep balance and avoid falling \cite{lapeyre2013thigh}, or have modified its open-source design into non-standard form factors to achieve more reliable walking behavior \cite{song2015development,popov2017design,teja2015optimal,lapeyre2014rapid,duminy2016strategic}. Nevertheless, the low cost and off-the-shelf availability of Poppy's standard design make it an attractive platform for researchers and educators, especially those whose expertise is oriented more towards software than hardware.  The many limitations of the standard design also make it useful for research specifically in resource- and compute-constrained robotics applications.  Therefore, it is desirable to have autonomous walking capability on the standard Poppy humanoid.


This paper delivers that capability, through a combination of closed-loop optimal control, and machine learning with previously collected, open-loop hardware runs.  The controller uses a Linear-Quadratic Regulator (LQR) to continually drive the system towards the nominal open-loop trajectory.  The linearized dynamics around the nominal trajectory, and a quadratic cost function that penalizes deviations more closely associated with falls, are both learned from the previously collected open-loop data using a novel convex optimization formulation.  Through extensive hardware validation, we demonstrate that the closed-loop controller achieves a statistically significant improvement in walking performance.  All data\footnote{{\scriptsize\href{https://drive.google.com/drive/folders/1sHA6S9qQq7ws1Z5cTDdqdGFlT0KdZI0Q?usp=sharing}{https://drive.google.com/drive/folders/1sHA6S9qQq7ws1Z5cTDdqdGFlT0KdZI0Q?usp=sharing}}} and code\footnote{\scriptsize\href{https://github.com/garrettkatz/poppy-muffin}{https://github.com/garrettkatz/poppy-muffin}} needed to reproduce our results and deploy the controller are open-source and freely available online.

\section{Background}

\subsection{Details of the Poppy Humanoid Platform}

The Poppy Humanoid is a child-sized humanoid robot, roughly 3.5kg in weight and 83cm tall.  It has 3D-printed plastic limbs, an RGB camera and Odroid UX4 embedded chip in its head running Ubuntu Linux, and 25 articulated joints, each with a position-controlled Dynamixel servo motor.  The Poppy Project also provides the PyPot Python API for sending target angles to the motors and retrieving their sensor readings, which include measurements such as temperature and voltage in addition to current joint angles.

Due to hardware and software limitations, torque control is not available and the control frequency is limited.  While sophisticated humanoid control often relies on frequencies over 100Hz, the Poppy platform is known to have maximum control rates that are substantially lower.  On our own Poppy hardware, motion became choppy and unreliable when sending target positions more frequently than 5Hz.  On the other hand, we could reliably read joint measurements at a higher frequency near 100Hz.  This means that our control loop can collect about 20 joint observations before issuing each target position command.

\subsection{Bipedal Locomotion Methods}

Bipedal robotic locomotion is a long-standing and active research area \cite{ha2025learning}.  In biological organisms, balanced and smooth locomotion is accomplished by the cooperation of the skeleton, joints and muscles. However, there are many issues to resolve when controlling bipedal robots to move in a stable fashion. Compared to quadruped robots, fewer legs result in inherent instability when physically interacting with the environment. Additionally, the system can more easily become underactuated when full contact is not preserved between the foot soles and the ground, or when the ankle joints do not have two full degrees of freedom.

Traditionally, bipedal locomotion research has focused on physics-based modeling. Vukobratovi\'{c} et al. \cite{vukobratovic1972stability} developed the well-known Zero Moment Point (ZMP) framework for stable walking pattern generation. The ZMP is the point inside the feet contact surface where the resultant force produces no horizontal moment. Although ZMP usually requires a simplified dynamics model (e.g., the 3D linear inverted pendulum \cite{kajita20013d} or double inverted pendulum \cite{caux1998balance}), it has been successfully applied to stable walking pattern generation on real robots such as HRP-2 \cite{kajita2003biped} and the AR-601M \cite{danilov2016zmp} robot. More generally, trajectory optimization \cite{kelly2017introduction} encodes the time-discretized equations of motion as constraints in a non-linear optimization problem.  

Walking is a periodic movement alternating double stance phases and single stance phases, depending on the number of feet contacting the ground. Contact events introduce discontinuities into the dynamics, so modeling the phases and corresponding transitions increases robustness of locomotion and makes it possible to handle various gaits. For example, Caron et al. \cite{caron2017make} designed a finite state machine for representing the single and double stance phases with the transitions triggered by geometric conditions.

Many recent works on bipedal locomotion have incorporated deep learning.  For example, a two-stage learning based model is proposed in \cite{he2025attention} for generalized legged locomotion. A convolutional neural network (CNN) embeds local terrain features in a robot-centric height map while another multihead attention module queries point-wise map features and combines them with proprioceptive observations. HugWBC \cite{xue2025unified} is designed for generating locomotion on humanoid robots with dynamic, customizable control, enabling the robot to perform gaits such as walking, standing, jumping, and hopping. A mathematical constraint inspired by morphological symmetry is introduced to encourage the policy to generate natural and symmetric motion, and extends to different reinforcement learning (RL) algorithms. Chen et al. \cite{chen2024reinforcement} formulate a model-based RL problem to learn a reduced-order model (ROM). The  control policy used on the robot is taken into account while optimizing the ROM, which effectively closes the performance gap between offline model optimization and online model deployment. Duan et al. \cite{duan2024learning} develop a sim-to-real learning approach for vision-based bipedal locomotion over challenging terrain. The model is composed of two learned components: a control policy that takes proprioceptive information and heightmap of a local region as inputs, and a heightmap predictor. The main contribution is the success of transferring the locomotion controller from simulation-only data to the real world.

\subsection{Linear Quadratic Regulators}


The Linear Quadratic Regulator (LQR) is an optimal control technique for feedback control of linear systems. Its objective is to stabilize a system while minimizing a quadratic cost function that balances performance and control effect. Although classical LQR applies to linear dynamics, many prior works have extended it to robotics where system dynamics are often strongly nonlinear. Li et al. \cite{li2004iterative} introduced the iterative LQR algorithm, which iteratively optimizes a target trajectory, and linearizes the system dynamics around the current target trajectory in each iteration. Tedrake et al. \cite{tedrake2010lqr} proposed LQR-Trees, integrating time-varying LQR with Lyapunov functions and sums-of-squares methods to generate feedback motion planners with formal stability guarantees. Kuindersma et al. \cite{kuindersma2014efficiently} developed a computationally efficient quadratic programming framework which leverages time-varying LQR design for ZMP dynamics, and demonstrated its effectiveness across a range of humanoid walking tasks. 



\subsection{System Identification}

Classical LQR assumes the linear dynamics model is already known, but in practice, it must often be refined or estimated based on data.  System identification (SI) is the process of creating a mathematical model to match the observed input-output behavior of a dynamical system. Early research in robotics primarily focused on estimating physical parameters with mathematical models. Atkeson et al. \cite{atkeson1986estimation} formulated robot inverse dynamics as a regression problem and estimated the inertial parameters for manipulator load and links. Traversaro et al. \cite{traversaro2013inertial} extended the regression framework to include the inertial, friction and motor parameters, and validated the identified model on the iCub humanoid robot \cite{metta2010icub}. 

In the context of reinforcement learning, SI has been increasingly adopted to regularize policy learning and bridge the sim-to-real gap. Zhu et al. \cite{zhu2017fast} proposed a Bayesian optimization framework that iteratively tunes simulator physical parameters during policy training, seeking parameters that are accurate enough to approximate the value of a locally optimal policy. Tan et al. \cite{tan2018sim} addressed sim-to-real transfer by improving simulation fidelity via SI, and learned robust controllers for the Minitaur robot. More recently, Memmel et al. \cite{memmel2024asid} proposed a learning framework that leverages limited real-world data to autonomously refine simulation parameters, enabling the learned controller to be reliably deployed in the real world. These works highlight how SI has evolved from a purely mathematical problem into a tightly integrated component of modern control frameworks.  Our work also involves a basic form of system identification to estimate locally linear dynamics models from data.



\section{Methodology}

Our approach starts with design of a nominal, imperfect walking trajectory that results in a mix of falls and successful runs during open-loop playback on the hardware.  We collect joint angle observations during these runs and manually label them according to how many footsteps were successful before a fall.  The collected data is then used to fit a linear dynamics model and quadratic cost function for a closed-loop LQR controller in the neighborhood of the nominal trajectory.  The details of this process are covered in the following subsections.

\subsection{Nominal Trajectory Design}

We designed a nominal walking trajectory through a combination of kinematic analysis, trajectory optimization, and manual tuning. First, we employed standard inverse kinematics techniques to solve for a trajectory that avoided swing foot collisions with the ground, kept the putative center of mass above the support polygon at all times, and moved slowly enough that dynamics effects were negligible.  However, this approach required the movement of the robot to be prohibitively slow.

Building off the initial design, we next explored trajectory optimization with physics-based dynamical constraints \cite{kelly2017introduction} to achieve a faster gait while maintaining balance.  However, the trajectories optimized in simulation did not transfer well to the physical hardware -- possibly because of system identification and modeling errors, or potentially due to low-frequency position control as opposed to high-frequency torque control.

Ultimately, using the previous designs as a starting point, we manually tuned the timing and target joint angles of each trajectory waypoint through empirical trial and error.  Through this manual tuning we were able to identify a walking trajectory that worked reasonably well in open-loop hardware playback.  The trajectory uses all leg motors and a subset of torso and arm motors; the others are kept fixed.  For reproducibility, the precise timing and target values for each joint are available in our code base.

\begin{figure}[t]
\centering
\includegraphics[width=\textwidth]{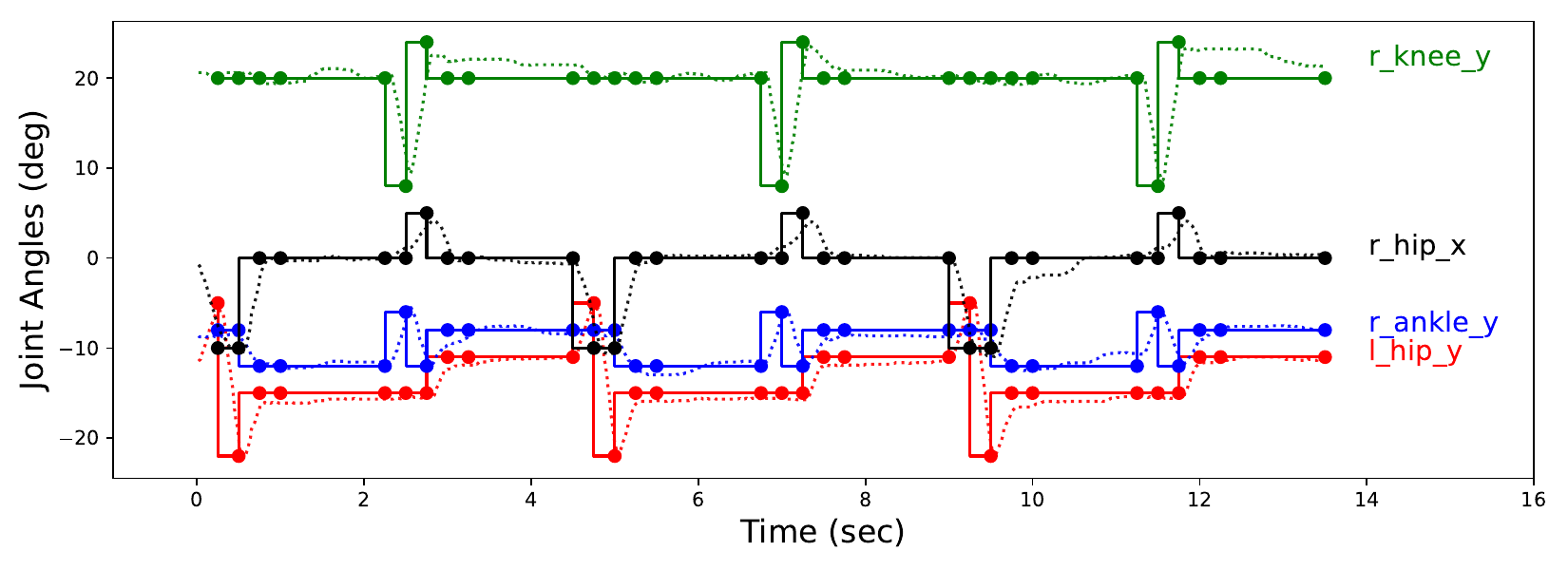}
\caption{Nominal trajectory (solid lines) and recorded joint angles (dashed lines) from a representative hardware run.  Circles indicate the timing of the waypoint commands.  Only a subset of joint motors are shown for figure legibility, labeled at right according to the standard Poppy naming convention.\label{fig:nominal}}
\end{figure}

The final nominal trajectory of target angles is visualized in Figure \ref{fig:nominal}, along with actually observed angles from a representative hardware run that did not include a fall.  This figure shows three cycles of the periodic gait, where each cycle consists of a left and right footstep.  During each swing phase, the waypoint command timing is close to the maximum 5Hz control rate; once both feet are planted there is a longer delay of 1.25 seconds before the next footstep to help stabilize the motor positions.

\subsection{Trajectory Playback}

When sending waypoint commands to the Dynamixel hardware, we used the low-level \texttt{DxlIO} interface in the PyPot API to set goal positions.  We kept most configuration settings at their defaults, with the following exceptions.  First, we increased the proportional gain of the motors' built-in PID position controllers from 4 to 8, which we found empirically necessary for motors to consistently reach their targets.  However, this also resulted in motors reaching their targets ahead of schedule.  To correct this effect, we commanded the motor to a goal position farther away than the intended target, but also capped its maximum speed accordingly, so that, if moving at that speed, the motor would reach the intended position at the intended time.  We applied this correction at all joint targets in the trajectory except the extrema, i.e., where the motor would be reversing direction for its subsequent waypoint.

\subsection{Open-Loop Data Collection}

\begin{figure}[t]
\centering
\includegraphics[width=\textwidth]{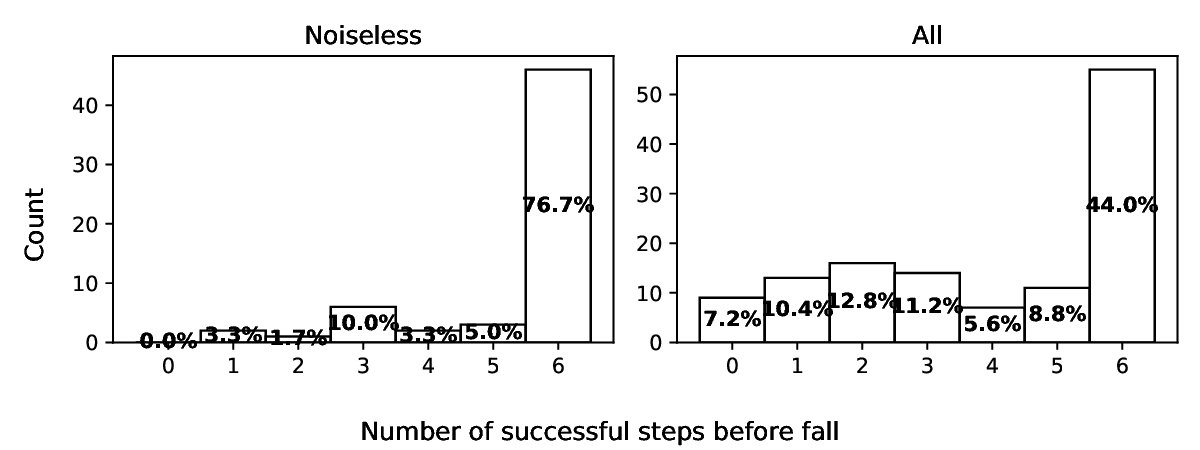}
\caption{Histogram over open-loop runs of the number of successful footsteps before a fall (trajectories were 6 footsteps long, so 6 successful footsteps means no fall during the episode).  The left panel is for noiseless trajectories only ($\sigma=0.0$, see text); the right panel is over all runs. \label{fig:controllability}}
\end{figure}

Using the trajectory and playback method described above, we collected a dataset of 125 open-loop hardware runs over a period from March to May in 2025.  Each run executed six footsteps of an open-loop trajectory, and the number of successful footsteps before a fall was recorded (six successful footsteps means no fall occurred).   These runs were collected in three locations, namely, all except the rightmost location pictured in Figure \ref{fig:poppy}.  The rightmost location was used later to assess generalization ability of the LQR-augmented control. To explore the dynamics in the neighborhood of the nominal trajectory, we added independent, zero-centered Gaussian perturbations to each target angle in each waypoint -- except the initial stance pose, which we never perturbed, so that suboptimal movements in one cycle would not have undue influence on subsequent cycles.  The standard deviation $\sigma$ of the perturbations was varied between $0.0$, $0.125$, and $0.25$ degrees across runs (but kept constant within any given run).  The $\sigma=0.0$ case was used to capture inherent variability in the system even when the exact same open-loop trajectory was provided.  The distributions of manually-labeled successful footstep counts, with and without perturbations, is shown in  Figure \ref{fig:controllability}.


\begin{figure}[t]
\centering
\includegraphics[width=\textwidth]{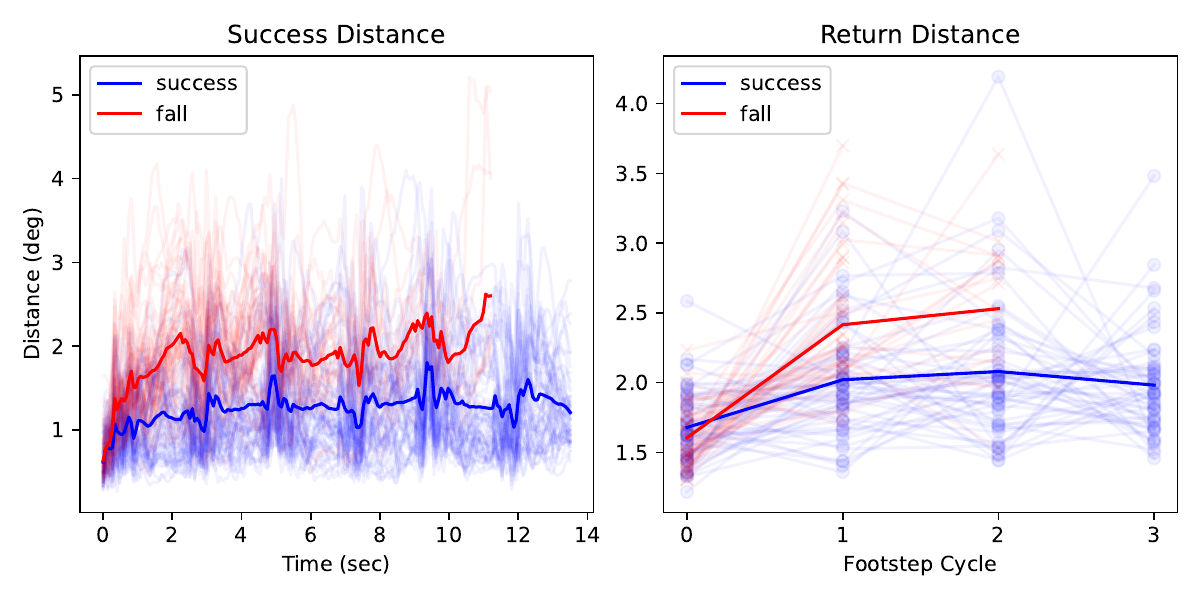}
\caption{Per-run, per-timepoint Euclidean distances from current joint observations to average successful run (left) or initial stance pose before each footstep cycle (right).  On right we only show the timepoints at the boundary of each cycle.  Each transparent curve is a separate run, and opaque curves show the means over all runs in their respective categories.  Data from failure runs after the recorded fall step are omitted from the plots.\label{fig:open_loop}}
\end{figure}

We also inspected the time evolution of the measured joint angles, and found evidence of gradual accumulation of errors, as expected when using open-loop control.  This effect is visualized in Figure \ref{fig:open_loop}.  The left panel shows Euclidean distance $||\theta_{r,t} - \hat{\theta}_{t}||_2$, where $\theta_{r,t}\in\mathbb{R}^{25}$ is the joint observation vector in run $r$ at time $t$, and $\hat{\theta}_t\in\mathbb{R}^{25}$ is the average observed joint vector at time $t$ across all successful runs.  The right panel shows $||\theta_{r,t} - \overline{\theta}_0||_2$, where $\overline{\theta}_0$ is the initial target stance pose at the beginning of each footstep cycle, and $t$ is limited to the timepoints in the data where a footstep cycle begins.  Both panels suggest that failure runs tend to involve joint observations with greater radial distance from some central point.  Therefore, it stands to reason that an LQR-style quadratic cost function should be sufficiently expressive to assign higher cost to observations that are more indicative of an impending fall.

\subsection{Residual Closed-Loop Trajectory Tracking with LQR}

We next employ LQR during execution to make small online adjustments to the target waypoints which will counteract accumulation of error. Formally, we will denote the $n$th target waypoint sent to the robot by $u_n\in\mathbb{R}^{25}$, a vector of intended joint angles for each of the $25$ joints.  The command is issued at time $t_n$ and the joint motors move towards their commanded targets over the time interval $[t_n,t_{n+1}]$, until the $(n+1)$th command is issued at time $t_{n+1}$.  During this time interval, actual joint angles are measured multiple times at a higher frequency, producing a sequence of measurement vectors $\theta_{\tau_1},...\theta_{\tau_M}$, with each $\theta_{\tau_i}\in\mathbb{R}^{25}$, and $t_n\leq \tau_1 \leq ... \leq \tau_M\leq t_{n+1}$.

The sequence length $M$ can vary due to noise and latency in the system.  To fit within the LQR formulation, this variable-length sequence must be converted to a fixed size and packaged as a single observation vector $x_n$.  We do this by linearly interpolating the measurements at a constant number $\overline{M}$ of equally spaced timepoints between $t_n$ and $t_{n+1}$, inclusive.  The interpolated measurements are concatenated into a single fixed-width state vector $x_{n+1}\in\mathbb{R}^{25\cdot\overline{M}}$.  In our experiments, we treat $\overline{M}$ as a hyperparameter that is varied across experimental conditions (but not varied within a run).  

Note that $x_n$ contains observations \textit{preceding} the command $u_n$, i.e. from the time interval $[t_{n-1},t_n]$.  This is because when each command $u_n$ is issued, it must be determined based on observations that have already been made.  At the edge case $x_0$ before any observations have been made, we use $\overline{M}$ copies of the initial joint measurements $\theta_0$.

The LQR controller is designed to drive the system towards a nominal trajectory $(\overline{x}, \overline{u})$.  It is therefore applied to the residuals $\delta x_n = x_n - \overline{x}_n$ and $\delta u_n = u_n - \overline{u}_n$.  We use a time-varying linearization of the system dynamics around the nominal trajectory, i.e.
\begin{align}
    \delta x_{n+1} \approx A_n \delta x_n + B_n \delta u_n.\label{eq:lindyn}
\end{align}
LQR determines an optimal linear control law that minimizes a quadratic cost function of the form
\begin{align}
    J(x,u) &= \delta x_N^\top Q_N \delta x_N + \sum_{n=0}^{N-1}\delta x_n^\top Q_n \delta x_n + \delta u^\top_n R_n\delta u_n + 2\delta x_n^\top S_n \delta u_n \\
    &= \delta x_N^\top Q_N \delta x_N + \sum_{n=0}^{N-1}[\delta x_n^\top, \delta u_n^\top] \left[\begin{array}{cc}Q_n & S_n \\ S_n^\top & R_n\end{array}\right]\left[\begin{array}{c}\delta x_n \\ \delta u_n\end{array}\right].
\end{align}
We estimate the dynamics and cost coefficient matrices, $A_n,B_n,Q_n,R_n,S_n$, from the open-loop data as described in the following sections.

The optimal linear control law has the form $\delta u_n^* = K_n\delta x_n$, where $K_n$ is computed recursively according to the well-known LQR equations
\begin{align}
    P_N &= Q_N \\
    K_n &= -(R_n + B_n^\top P_{n+1}B_n)^{-1}(B_n^\top P_{n+1}A_n + S_n^\top) \\
    P_n & = Q_n + A_n^\top P_{n+1}(A_n + B_nK_n) + S_nK_n
\end{align}
with intermediate matrix variables $P_n$.

\subsection{Linear System Identification}

We estimate the linearized dynamics matrices $A_n$ and $B_n$ directly from the data with a least-squares linear fit.  Specifically, for each waypoint $n$, we pool the residuals $\delta x_n$ and $\delta u_n$ measured across all open-loop runs by arranging them as columns of matrices $\delta X_n$ and $\delta U_n$.  We exclude data from during and after failed footsteps, since at this point the strap is taut and the human experimenter may have intervened to minimize wear and tear on the hardware, which means the measurements are no longer representative of autonomous system dynamics.  Using the remaining data, we recover $A_n$ and $B_n$ by solving for the block matrix $[A_n, B_n]$ in
\begin{align}
    \delta X_{n+1} = \left[\begin{array}{cc}A_n & B_n\end{array}\right]\left[\begin{array}{c}\delta X_n \\ \delta U_n\end{array}\right].
\end{align}
For sufficiently small $\overline{M}$, this is an overdetermined system and hence overfitting is not a concern.  To further mitigate overfitting, we pool data across cycles of the periodic nominal trajectory.  Since there are 3 cycles, each consisting of 10 waypoints, this means that we fit 10 pairs $(A_0,B_0),...,(A_9,B_9)$, and each pair $(A_n,B_n)$ is fit using data from waypoints $n$, $n+10$, and $n+20$.

\begin{figure}[t]
\centering
\includegraphics[width=\textwidth]{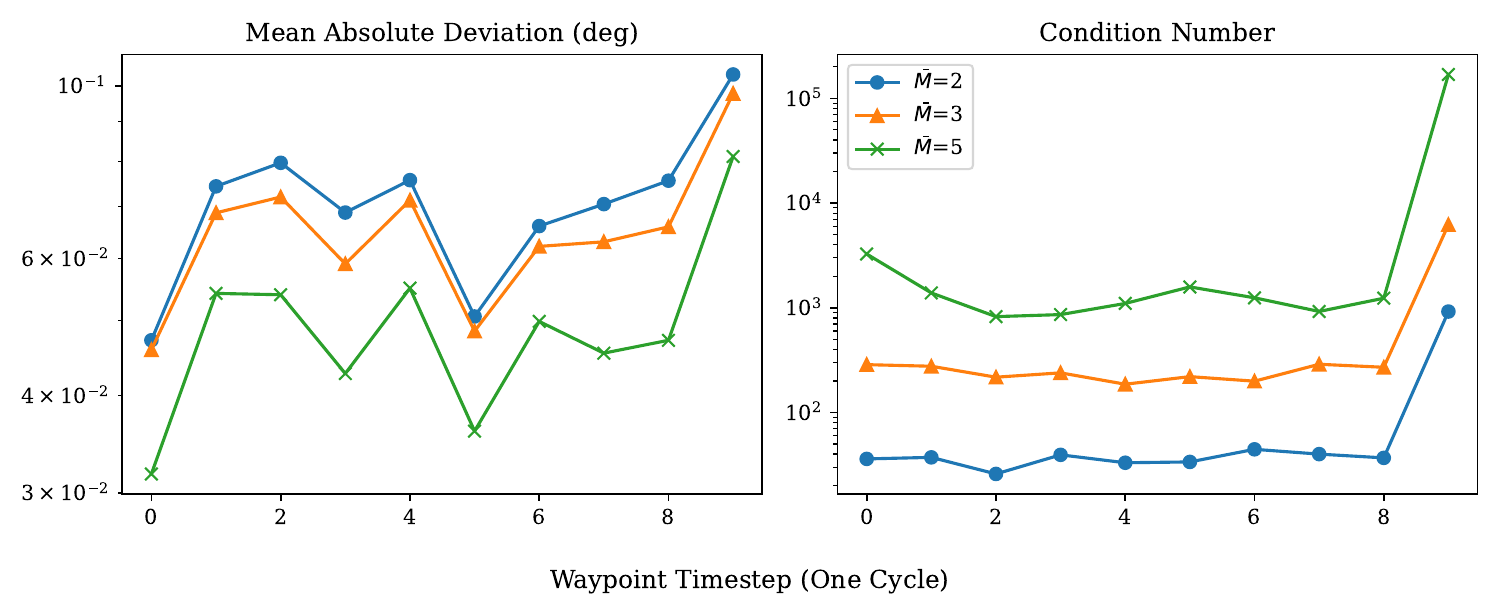}
\caption{\textbf{Left:} Mean absolute deviation between dynamics model predictions and observed data.  \textbf{Right:} Condition number of fitted dynamics matrix.\label{fig:dynamics}}
\end{figure}

Figure \ref{fig:dynamics} shows the results of the linear fits for each $n$ in one cycle. We compare different values for $\overline{M}$, the fixed number of interpolated timepoints when constructing the observation vector $x_n$. We measure the quality of the fit with the mean absolute deviation, i.e.
\begin{align}
    \text{MAD}_n = \text{mean}_{r,j}|\hat{\delta x}_{r,n+1,j} - \delta x_{r,n+1,j}|,
\end{align}
where $j$ indexes the joints and $\hat{\delta x}_{r,n+1} = A_n\delta x_{r,n} + B_n\delta u_{r,n}$ is the next state observation predicted by the fitted model in run $r$.  We also report the condition number of the fitted matrix $[A_n,B_n]$.

The results show that for all $\overline{M}$, the average absolute error in any predicted joint angle is around $0.1^\circ$ or less.  However, the condition number increases substantially with increasing $\overline{M}$ -- possibly because larger $\overline{M}$ increases the dimensionality of the datapoints relative to the number of datapoints, thereby reducing the regularizing effect of under-parameterization. Since large condition numbers could introduce numerical instability or large inaccurate dynamics predictions later in the process, we limited our remaining experiments to $\overline{M}=2$.  Smaller $\overline{M}$ also alleviates the computational burden of the quadratic cost fitting step, described next.

\subsection{Learned Cost Function}

Whereas the dynamics are dictated by the physical system, we have more flexibility in determining the cost function.  We have two primary considerations in our cost design.  First, we want the resulting controller to avoid falls as much as possible.  Second, we want the resulting controller to remain close to the nominal trajectory -- this prevents trivial solutions such as standing in place to avoid falling.

We combine these considerations into a constrained optimization problem.  The objective is to make the block coefficient matrices as close as possible to the identity, which will encourage trajectories that are close to nominal:
\begin{align}
    \underset{C}{\min} \sum_n \left|\left|C_n - I\right|\right|_\text{Fro}^2, \quad C_n = \left[\begin{array}{cc}Q_n & S_n \\ S_n^\top & R_n\end{array}\right]
\end{align}
Next, we impose constraints to ensure that fall trajectories in the data are assigned higher average cost per unit time than successful ones:
\begin{align}
    \forall(x,u)\in\mathcal{F}:\quad & \frac{1}{n_F}J(x_{\leq n_F}, u_{\leq n_F}) \geq d + \epsilon \\
    \forall(x,u)\in\mathcal{S}:\quad & \frac{1}{N}J(x, u) \leq d - \epsilon
\end{align}
where $\mathcal{F},\mathcal{S}$ are the sets of fall and success runs, respectively, $n_F$ is the first waypoint of a footstep where a fall occurs, and $(x_{\leq n_F},u_{\leq n_F})$ denotes the leading trajectory up until waypoint $n_F$.  For successful runs, $N$ is the maximal number of waypoints executed, namely 30 (five per footstep).  Here $d\in\mathbb{R}$ is an auxiliary scalar optimization variable that acts like a decision boundary, and $\epsilon$ is a hyperparameter specifying the margin required for the constraints.  A feasible solution for $d$ with $\epsilon > 0$ is equivalent to the requirement that all fall runs have strictly higher cost than successful runs.  We can also set $\epsilon \leq 0$ to allow some limited amount of constraint violation.

This optimization problem is in fact convex, since the constraints are linear with respect to the coefficient matrices and the objective is a convex quadratic.  To meet the requirements of LQR, we also need each $C_n$ to be positive semi-definite, but this is also a convex constraint.  Hence this optimization problem is a semi-definite program, and we can fit an LQR cost function to our data via convex optimization, which we implement with the CVXPY library \cite{agrawal2018rewriting}.  Our formulation has the added benefit of dealing with credit assignment implicitly, since it will automatically distribute the cost among the waypoints in a way that distinguishes stands from falls.

On the other hand, one limitation in the present approach is the assumption that in fall episodes, the ``point of no return'' (when falling becomes unavoidable) is no later than $n_F$.  Our hand is forced in this regard, because manual fall labeling does not allow finer temporal resolution than the footstep level.  Consequently, the precise fall time after $n_F$ is unknown -- and as soon as a fall begins, the observed joints will no longer determine the full system state (once the stance foot is not firmly planted, the base frame position and orientation is unknown).  Nevertheless, this assumption is borne out empirically by our experimental validation. 

\section{Empirical Validation}
\label{sec:results}

We validated the foregoing methodology in two stages. First, we tuned the hyperparameter $\epsilon$ based on stability analysis of the resulting LQR controller.  Second, we performed a large-scale controlled hardware experiment, empirically comparing the closed-loop LQR control with open-loop trajectory playback.  These two stages are detailed in the following subsections.

\subsection{LQR Stability Analysis}

Once cost functions are fit and we compute the LQR solutions $K_n$, note that the closed-loop dynamics become
\begin{align}
    \delta x_{n+1} &= A_n \delta x_n + B_n \delta u_n^* \\
    &= A_n \delta x_n + B_n K_n \delta x_n \\
    &= (A_n + B_n K_n)\delta x_n.
\end{align}
Therefore, the stability of the controller (whether it eventually drives deviations from the nominal trajectory towards zero) is measured by the maximum eigenvalue magnitude $\lambda^*_n$ of the matrices $A_n + B_n K_n$. Ideally $\lambda^*_n<1$ for all $n$, but some transient mildly expansive behavior may be acceptable as long as the cumulative products $\Lambda_n = \prod_{n'\leq n}\lambda^*_{n'}$ eventually stay below 1 by the end of one period of the gait.

\begin{figure}[t]
\centering
\includegraphics[width=\textwidth]{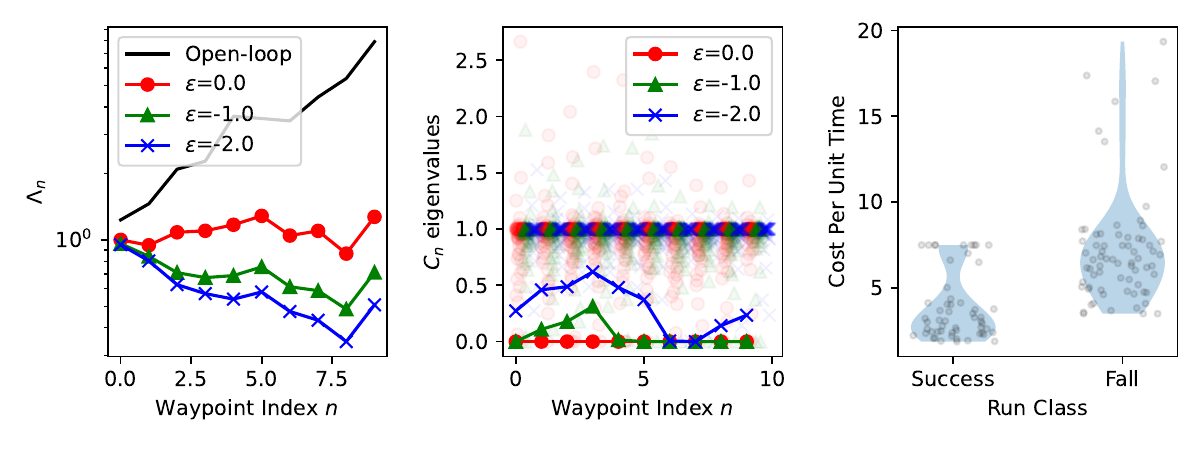}
\caption{\textbf{Left}: Cumulative products $\Lambda_n$ of maximum eigenvalue magnitudes of $A_n + B_nK_N$, for several $\epsilon$.  ``Open-loop'' refers to ``passive'' residual dynamics with no LQR adjustments, i.e., using maximum eigenvalues of $A_n$.  \textbf{Middle:} Eigenvalues of the cost coefficient matrices $C_n$.  Each scatterpoint is an individual eigenvalue and solid lines show minimum eigenvalue for each $\epsilon$ and $n$.  \textbf{Right:} Violin plots showing cost distributions over failed and successful runs in the open-loop data for $\epsilon=-2.0$ (each scatterpoint is a run with horizontal noise added for legibility). \label{fig:stability}}
\end{figure}

These cumulative products are shown in Figure \ref{fig:stability} (left) for different values of the cost-fitting hyperparameter $\epsilon$.  Recall that $\epsilon \leq 0$ allows some limited violation of the constraint that every fall run be assigned higher cost than every success run.  We observe that $\epsilon = 0$ produces an unstable controller, but $\epsilon < 0$ can achieve stability.

We can explain this effect by inspecting the eigenspectra of the cost coefficient matrices $C_n$ (Figure \ref{fig:stability}, middle).  The plot shows that for $\epsilon$ closer to zero, more $C_n$ matrices tend to be singular (having a zero eigenvalue).  Geometrically, this means the cost function is more like a parabolic valley than a bowl, since the corresponding zero eigenvector is a direction in which cost does not increase.  Apparently, the level sets of the cost function need to be flatter in certain directions to better separate failed and successful runs, since $\epsilon$ closer to zero means stricter enforcement of the cost separation constraints.

If a direction is not penalized by the cost function, then LQR will not be incentivized to squash residual trajectory deviations in that direction.  Therefore, the singular $C_n$ issue can translate to unstable directions in the LQR controller.  Since stable control is paramount, we selected the $\epsilon=-2$ version of the controller for hardware validation.  Despite allowing some violation of the constraints, the fitted cost function still tends to assign higher cost to fall runs.  This is shown in Figure \ref{fig:stability} (right), where the cost overlap between failure and success is limited to the band of $\pm\epsilon$ around the analogue of a ``decision boundary'' $d\approx 5$.

\subsection{Closed-Loop Hardware Performance}

Having selected our controller, we performed a large-scale experiment to compare walking performance using closed-loop LQR control vs.\@ open-loop playback of the original trajectory.  The performance metric was the number of successful footsteps before a fall.  Our null hypothesis was that performance in each case would follow the same distribution, with the alternative hypothesis that the LQR controller performance would be distributed more highly.

We repeated many runs in each condition (closed- vs.\@ open-loop) to ensure reproducibility and and collect enough measurements for a statistically significant result.  Operating within the time constraints of our team, we were able to collect 100 runs for each condition in the office location (leftmost panel of Figure \ref{fig:poppy}), and 80 runs for each condition in a lab location (rightmost panel of Figure \ref{fig:poppy}).  Note that the latter location had not been used when collecting training data, so it allows us to assess the generalization ability of the controller.

Runs were collected in batches of 20 with at least 5 minutes of rest for the motors in between each batch.  We alternated batches between closed- and open-loop control to counterbalance the experiment.  After data collection we employed a Mann-Whitney $U$ hypothesis test, since the successful footstep counts are ordinal but not normally distributed.  We used the SciPy \cite{2020SciPy-NMeth} implementation of the test with the ``asymptotic'' option since our sample size is large and continuity correction since the samples will contain many ties.

In the office environment, open-loop success rate was 62\% with an average of 4.18 steps before a fall, and closed-loop success rate was 78\% with an average of 5.13 steps before a fall ($p$ value of 0.00225).  Similarly, in the lab environment, closed-loop control improved the success rate from 30\% to 42.5\%, and average successful steps from 3.75 to 4.8 ($p$ value of 0.00014).  The overall performance drop in the lab environment could be explained by its different distribution from the training data and its very smooth floors relative to the office, which is carpeted.  We also observed that open-loop control at the time of this hardware validation was substantially worse than at the time of the original March-May 2025 data collection.  This may be due to accumulated wear and tear on the hardware, resulting from additional demonstrations and education usage in the time that had passed.  Regardless, our empirical validation clearly shows that our closed-loop control method exhibits a statistically significant performance improvement over open-loop playback, in both testing locations.

\section{Conclusion}

This work constitutes a first step towards reliable and versatile bipedal locomotion on the standard off-the-shelf Poppy Humanoid.  We have presented a methodology for closed-loop locomotion based on LQR control and quadratic cost functions that are learned from data.  Our empirical validation demonstrates that this methodology achieves statistically significant improvement in walking performance.

That said, several important directions remain for future research.  First, our system identification used a simple linear fit to the data.  There are more sophisticated SI methods in the literature which could be integrated into our procedure.  Second, there are additional sensor measurements available in the open-loop data which could potentially be useful for better control, including joint temperatures and loads as well as egocentric RGB video from the head camera.

Finally, although our controller can take steps and avoid falls, we have not yet optimized other important gait properties, such as forward velocity, turning radius, or cost of transport. This is due in part to difficulty in measuring these properties, but future work could measure them directly with additional experimental tooling, or estimate them indirectly from the existing sensors -- for instance, using optical flow analysis on the egocentric video feed.  Once these properties can be measured or estimated, they can potentially be optimized: for example, by incorporating goal-conditioned reinforcement learning with different target forward velocities as goals, or control-theoretic analogues thereof.

\bibliographystyle{splncs04}
\bibliography{main.bib}
%






\end{document}